\documentclass{article}
\usepackage{spconf,amsmath,graphicx}

\usepackage{makecell}
\usepackage{arydshln}

\title{Retrieval Augmented End-to-End Spoken Dialog Models}
\name{
\begin{tabular}{cc}
Mingqiu Wang, Izhak Shafran, Hagen Soltau, Wei Han, Yuan Cao, Dian Yu, Laurent El Shafey \\
\{mingqiuwang, izhak, soltau, yuancao, dianyu, shafey\}@google.com
\end{tabular}
}

\address{Google DeepMind}

\begin{document}

\maketitle

\begin{abstract}
We recently developed SLM~\cite{SLM2023}, a joint speech and language model, which fuses a pretrained foundational speech model and a large language model (LLM), while preserving the in-context learning capability intrinsic to the pretrained LLM. In this paper, we apply SLM to speech dialog applications where the dialog states are inferred directly from the audio signal. 

Task-oriented dialogs often contain domain-specific entities, i.e., restaurants, hotels, train stations, and city names, which are difficult to recognize, however, critical for the downstream applications. Inspired by the RAG (retrieval-augmented generation) paradigm, we propose a retrieval augmented SLM (ReSLM) that overcomes this weakness. We first train a speech retriever to retrieve text entities mentioned in the audio. The retrieved entities are then added as text inputs to the underlying SLM to bias model predictions. We evaluated ReSLM on speech MultiWoz task (DSTC-11 challenge), and found that this retrieval augmentation boosts model performance, achieving joint goal accuracy (38.6\% vs 32.7\%), slot error rate (20.6\% vs 24.8\%) and ASR word error rate (5.5\% vs 6.7\%). While demonstrated on dialog state tracking, our approach is broadly applicable to other speech tasks requiring contextual information or domain-specific entities, such as contextual ASR with biasing capability.

\end{abstract}

\section{Introduction}

Retrieval-augmented generation (RAG) models~\cite{lewis-etal-2020-retrieval, guu-etal-2020-retrieval, karpukhin-etal-2020-dense, borgeaud-etal-2021-improving, izacard-etal-2022-atlas} are increasingly seen as a valuable enhancement for LLMs in many natural language processing (NLP) tasks. Retrieval augments a LLM's ability to deliver accurate, grounded responses, especially for queries demanding specialized domain knowledge. For example, customer service chatbots can leverage personalized data and service manuals for more relevant response thus enhanced user satisfaction.

The same idea can be borrowed from the text domain to the speech domain to address the challenge of identifying rare domain-specific entities. Most traditional speech dialog systems rely on a cascaded approach where an automatic speech recognition (ASR) module first transcribes the audio input into text, and then the resulting transcript is fed into a downstream natural language understanding (NLU) system~\cite{zhao2022description}. This approach suffers from two key limitations: (1) it lacks the ability to correct potential ASR errors, and (2) both ASR and NLU modules struggle with entities poorly represented in their training data. For example, an ASR system might misspell ``Bermondsey", a London train station, as ``Bermondsay" due to local pronunciation. However, in a real-world train station reservation system, these London train stations and their correct spellings are readily available as additional contextual information, yet are underutilized in many existing systems. We hypothesize that using a retrieval module to integrate the additional contextual information into speech dialogue systems can provide correct spellings, mitigating the impact of ASR errors, thus enhancing the speech dialog performance.

Our paper addresses these challenges using DSTC-11 dialog tracking task~\cite{DSTC11} as the benchmark. This task, based on the MultiWoz corpus of human-human conversations~\cite{multiwoz21}, replaces written user responses with spoken utterances collected from crowdsourced workers. The model aims to infer the dialog state when conditioned on the current user utterance and the dialog history. The DSTC-11 task is particularly relevant due to the high frequency of domain-specific entities like restaurants, tourist attractions, cities, and train stations. Our paper makes the following contributions:
\vspace{-.04in}

\begin{itemize}
\itemsep 0in
\item {\bf Speech retriever component}: We introduce a speech retriever based on the dual-encoder architecture~\cite{ni2021large} where audio embeddings guide the retrieval of relevant text entities. 
\item {\bf Simple retrieval integration}: We propose a straightforward yet effective method for retrieval integration. By concatenating the retrieved entities to the LLM inputs, we provide the model with additional context to improve its ability to handle rare entities.
\item {\bf Performance gains on DSTC-11}: Our direct audio-to-dialogue-state approach outperforms all the submissions in the DSTC-11 challenge~\cite{DSTC11}, where all of them used cascaded ASR + NLU systems. We show that the use of retrieval augmentation led to marked gains compared to the baseline.
\end{itemize}

Although we experiment with this approach on the dialog state tracking task, it has broader applications in various speech understanding tasks that use domain-specific languages, such as contextual ASR.
\section{Related work}





Lately, retrieval-augmented language models have demonstrated superior performance on various natural language tasks~\cite{lewis-etal-2020-retrieval, guu-etal-2020-retrieval, karpukhin-etal-2020-dense, borgeaud-etal-2021-improving, izacard-etal-2022-atlas, gao2023retrieval}, especially for knowledge-intensive ones. This approach disentangles domain-specific knowledge from training parameters of LLMs, partially providing a potential solution to the LLM hallucination issue~\cite{huang2023survey}. Contrasted with the widespread adoption of RAG in the NLP field, its application in the multimodal domain is less prevalent. RA-CM3~\cite{yasunaga2023retrieval} first presented a multimodal model of both retrieving and generating text and images. For audio modality, Chan et al~\cite{chan2023using} used kNN based selection to leverage external knowledge in ASR system. Yuan et al~\cite{yuan2023retrieval} applied RAG to audio generation.

Domain-specific speech recognition and spoken language understanding tasks are challenging partially because rare entities are generally difficult to be transcribed or understood by ASR or NLU systems. Notably, additional contextual information is proved to benefit ASR error correction and downstream NLU systems~\cite{lin2015hierarchical,zhao2019shallow,liu2017dialog,jonson2006dialogue, raju2018contextual,jaech2018personalized,kim2018dialog,williams2018contextual}. Contextual information from different sources can be quite lengthy and noisy, presenting a open-ended research in its effective utilization. In this work, we explore the feasibility of apply RAG in speech understanding field. Domain knowledge is crucial for speech task-oriented dialog task~\cite{wu-etal-2019-transferable, zhou-small-2019-multi}. Additionally adaptation to new schemas might be required for unseen domains~\cite{rastogi-etal-2020-towards}. Therefore we use speech dialog understanding task as a testing ground to evaluate our approach.


\section{Model}

\subsection{Joint speech and language model (SLM)} \label{sec:slm}

\begin{figure}[ht]
    \centering
    \includegraphics[width=\columnwidth]{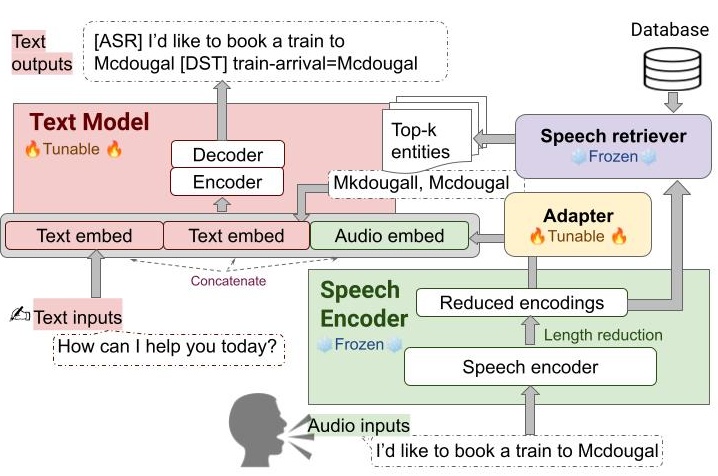}
    \caption{ReSLM architecture: a speech retriever (trained separately) is added onto SLM, which can provide additional text input as contextual information to SLM. Using the reduced encodings from the speech encoder as queries, the retriever finds the top-k nearest entities (text) and the entities are concatenated to the text inputs entering SLM.}
    \label{fig:reslm}
\end{figure}

Speech dialog understanding requires a model that can simultaneously recognize speech contents\footnote{Does not necessarily require verbatim transcription.} and understand the semantic meaning behind the spoken language. Furthermore, the model needs to support both speech and text inputs to deal with bimodal dialog history, where user turns and agent turns can be possibly in different modalities (audio or text). In the DSTC11 challenge, user turns are always in speech while agent turns are in text. We adopt the join speech and language model (SLM)~\cite{SLM2023} as our backbone, as shown in Figure~\ref{fig:reslm}.

SLM adapts a pretrained speech model to a pretrained large language model (LLM) using a simple adapter. The speech input is fed into the speech encoder (we used USM encoder~\cite{usm2023} in this work). The output speech embeddings are down-sampled to reduce the sequence length so as to allow longer speech input (See details in section 3.1 in ~\cite{SLM2023}). The reduced speech embeddings are then fed into the speech-to-text adapter (2 layers of Transformer~\cite{Vaswani2017} in this work), and the output from the adapter is concatenated along the time dimension with text input embedding sequence. The concatenated sequence then enters into the LLM Transformer stack (we used the T5-FLAN~\cite{chung2022scaling, raffel2020exploring} in this work). The output of LLM is the dialog state string to be predicted. We use the next-token-prediction loss as the training objective.



\subsection{Retrieval-augmented SLM (ReSLM)} \label{sec:reslm}

In short, ReSLM is basically a SLM with a retriever add-on. To utilize the additional contextual information, we first train a speech retriever to rank the contextual information.
The goal of the retriever is to retrieve a set of entities (train stations, hotel names, etc. in DSTC11 task) that are likely to be mentioned in the speech utterance. In this way, we re-frame the problem from free text generation (generating dialog state strings containing rare entities) into a ranking problem, where the retriever helps with ranking a list of potential mentions. This helps reduce the task difficulty.

The retriever is based on the `dual encoder' architecture~\cite{ni2021large}, where one encoder encodes the query and the other encoder encodes the candidate into two fixed-length embeddings. For simplicity, our implementation adopts a shared encoder for both query (audio of user's utterance in our case) and candidate (text of entity names in our case). An average pooling is applied on the encoder outputs (generally varied length) to map it to the fixed length embedding. The relevance of an audio / entity pair is scored by the cosine distance~\cite{ni2021large}. 

During training time, the input consists of audio of user utterances and entity names mentioned in the reference transcript for creating positive pairs. Note that, one utterance can mention multiple or zero entities. The in-batch negatives are used for creating negative pairs. The model is optimized using the contrastive loss. During inference, we find the nearest neighbors efficiently using cosine distance with the SCAM library and retrieve the top-K candidates~\cite{guo2020accelerating}.

To augment the retrieval results to the underlying SLM, we simply concatenate the retrieved entities to other text inputs entering SLM. For example, SLM takes the text input of {\em dialog history}; while ReSLM takes the text input of {\em dialog history, retrieved entities}. Note that SLM and the retriever are two separated models and we don't back-propagate gradient of SLM into the retriever.
:q

\section{Experiments and results}
\label{sec:results}

\subsection{The DSTC11 Challenge}
Building on the popular MultiWoz 2.1, dialog state tracking task~\cite{multiwoz21}, DSTC11 challenge replaces written user responses with spoken utterances, generated by TTS for the training set and human volunteers for the test set~\cite{DSTC11}. This spoken version focuses on speech dialog state tracking, with performance measured by Joint Goal Accuracy (JGA) and Slot Error Rate (SER) as outlined in the challenge details. Additionally, for ablation experiments analyzing the impact of misrecognitions, we also measure the Word Error Rate (WER) of the recognized transcripts of speech inputs.

\subsection{Auto-Regressive Inference} 
Dialogs have multiple turns and the dialog state values are inferred turn-by-turn auto-regressively. The task of dialog state tracking requires predicting all dialog states up to the current turn $i$, therefore the entire dialog history is required as input. For example, if a user mentions to book a flight to New York in the 1st turn, {\em flight-destination=New-York} keeps as a dialog state until the last turn unless this request is changed during the dialog.

\begin{figure}[ht]
    \centering
    \includegraphics[width=\columnwidth]{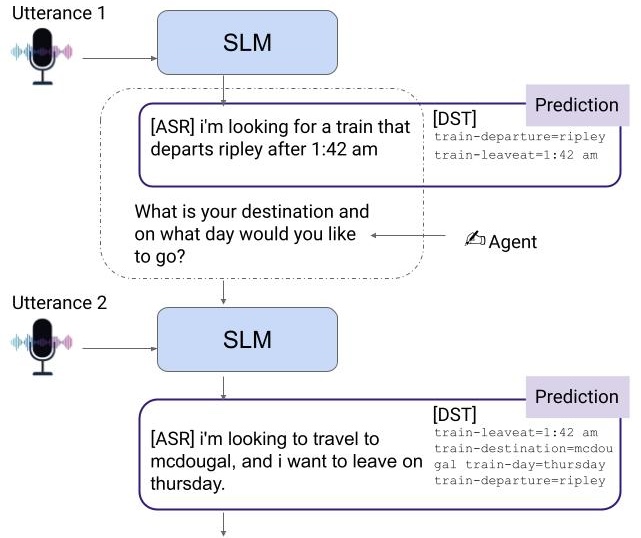}
    \caption{Direct speech to dialog state prediction in multi-turn dialogs.
    Given the speech of the current user turn $i$ and a text transcript of the dialog history, SLM or ReSLM generates a single output sequence for both the corresponding transcript {\em [ASR]} and dialog state {\em [DST]}. The ASR transcript predicted from the turn $i$ is used as history for the turn $i+1$ in an auto-regressive manner.}
    \label{fig:auto-regressive}
\end{figure}

As shown in Figure \ref{fig:auto-regressive}, we feed the speech of turn $i$ as audio input and the dialog history from turn $1$ to $i-1$ as text input. The dialog history can be long (up to 10-20 min) and was the best represented in the text form, not speech, without a suitable longform speech model available at the moment when this experiment was conducted. To obtain dialog history transcripts, we trained SLM to simultaneously recognize the words spoken in turn $i$ along with the dialog states in one output stream. The transcript from each turn is incrementally collated to create the dialog history for subsequent turns.

During the training time, the input consists of speech of the current turn and the dialog history transcribed by the previously described USM model. The loss is computed on targets consisting of the reference transcript of the current turn and the associated reference dialog state values.

During the inference time, the ASR transcript predicted by SLM/ReSLM from turn $i$ is used as history turn $i+1$ in an auto-regressive manner.

\subsection{Baseline (SLM)}

We use SLM (Section~\ref{sec:slm}) as our baseline, which combines a pretrained speech encoder, 2B USM ~\cite{usm2023}, with a pretrained LLM, T5-FLAN model of size XXL (13B)~\cite{chung2022scaling, raffel2020exploring}.

Same as described in the SLM paper~\cite{SLM2023}, we first freeze both the speech encoder and the LLM while only train the adapter. We use ASR task for the adapter training: given a user speech, predicting the corresponding transcript. After the adapter is pretrained, we finetune both the adapter and the encoder of T5 using the dialog state tracking task: given a user speech and the dialog history transcript, predicting the dialog state until this user turn. The reason for such a two-stage training is because ASR is a relatively easy task to warm up the adapter from scratch, and finetuning is needed because the original T5 model does not contain the 0-shot dialog state prediction capability. Note that, it is also possible to combine these two stages and directly train the adapter from scratch with T5 finetuned at the same time. But we found this 1-pass training is sub-optimal compared to the 2-passes: Before the adapter learns to map the audio encodings into the text embedding space, unfreezing T5 weights and tuning together with the adapter may leads to the capability forgetting of the pretrained T5.

The baseline model achieves 32.7\% JGA on inferring dialog states and 6.7\% WER on recognizing the user response (Table~\ref{tab:reslm_results}). We analyzed the common errors, as shown in Table~\ref{tab:ser_results}, and found that majority of errors come from mis-spelled entities, such as hotel, restaurant, and city names. To improve the domain-specific entity recognition, we added the retrieval component to SLM, which leverages the known database of available entities as additional contextual information.
\subsection{Retrieve Augmented Model (ReSLM)}
The speech-to-text retriever, described in Section~\ref{sec:reslm}, was trained on a curated set of 2,500 entities, including hotels, restaurants, and city names, extracted from the full DSTC11 challenge \cite{DSTC11} training set. While the retriever could utilize any audio and text encoders, we opted for the SLM encoder (speech encoder plus T5 encoder) for both queries (audio input) and values (text input), leveraging SLM's bimodal capability.

A separate pool consisting of different 14k entities was used for evaluation, where the 14k entities are curated from the full DSTC11 test set. Table~\ref{retriver} showcases the retriever's performance. We selected at most top-10 entities per utterance based on a distance threshold of -0.78 (we applied the intersection of them, resulting the actual entity number ranging around 4-5 per utterance). This selection strikes a balance between recall and precision. We prioritized recall to ensure the resulting ReSLM to have access to a wider range of potential entities, even if it compromised precision.

\begin{table}[ht]
  \centering
  \begin{tabular}{|c|c|c|} \hline
   Top-k & Recall & Precision \\ \hline
    1  & 40.2\%  & 13.3\% \\ \hline
    3  & 51.9\%  & 6.7\% \\ \hline
    5  & 57.0\%  & 5.0\% \\ \hline
    10  & 62.2\% & 3.6\% \\ \hline
    20  & 66.5\% & 2.8\%  \\ \hline
    100 & 70.4\% & 2.0\%  \\ \hline
  \end{tabular}
 \caption{Performance of the retriever. We selected top-10 entities with a distance threshold $-0.78$ for later ReSLM training and evaluation.}
\label{retriver}
\vspace{-.1in}
\end{table}

By augmenting the retrieved entities into SLM, ReSLM demonstrates a significant gain in dialogue state tracking (DST) performance. As shown in Table~\ref{tab:reslm_results}, JGA leaps from 32.7\% to 38.6\%. ReSLM outperforms all participants from the challenge where the best performing system achieved 37.9\% JGA~\cite{DSTC11}.

\begin{table}[ht]
\centering
\begin{tabular}{|l|c|c|} \hline
Model & JGA $\uparrow$ & WER $\downarrow$ \\ \hline
DSTC11 Cascaded/T5 (\cite{DSTC11}) & 30.9\% & 13.0\% \\
DSTC11 Whisper/T5 (\cite{DSTC11})  & 34.3\% & 8.9\% \\
Best DSTC11 system (\cite{DSTC11}) & 37.9\% & - \\ \hline
SLM                                & 32.7\% & 6.7\% \\
Retrieval augmented (ReSLM)        & {\bf 38.6}\% & {\bf 5.5}\% \\ \hline
\end{tabular}
\caption{Performance of JGA and WER on DSTC11 challenge. Adding a retrieval component improves the DST performance substantially, from 32.7\% to 38.6\% JGA.}
\label{tab:reslm_results}
\end{table}

\begin{table}[ht]
\centering
\begin{tabular}{|l|c|c|} \hline
Category & SLM  & ReSLM \\ \hline
hotel name & 42.1\% & 30.6\% \\
restaurant name  & 40.1\% & 30.9\% \\
train departure city & 60.5\% & 40.5\% \\
train arrival city & 63.1\% & 39.6\% \\ 
train departure time & 52.3\% & 46.1\% \\
train arrival time & 31.6\% & 22.4\% \\
\hline
\end{tabular}
\caption{Categorical slot error rate (SER). ReSLM outperforms SLM in entity names such as hotel, restaurant and departure/arrival cities.}
\label{tab:ser_results}
\vspace{-.1in}
\end{table}

Our ablation studies in Table 3 show that the gains come from recognizing entities that are specific to the task such as hotel names (27\% gain), restaurant names (23\% gain), train destination station (33\% gain), and train arrival stations (37\% gain). These are typically in the long-tail of the vocabulary and difficult to model by ASR systems. We also see gains in the extraction of time related information, though the extracted time information can be further improved using number normalization (e.g., "three pm" vs "3pm"). Not surprising, retrieval augmentation also improves the speech recognition of user utterances with a gain of 18\% WER.

\section{Conclusions}
We introduce a retrieval-augmented speech understanding model (ReSLM). ReSLM can leverage the given contextual information using a speech retriever and make more grounded thus accurate prediction. We experimented this idea in DSTC11 speech-aware dialog challenge. We showed that our direct audio-to-dialog-state approach using ReSLM significantly enhances the dialog state tracking performance, particularly for domain-specific entities like hotel, restaurant, and city names, by simply concatenating the retrieved entities to the text inputs fed to the underlying SLM. ReSLM outperforms all the submissions in the DSTC11 challenge, achieving 38.6\% JGA in dialog state tracking and 5.5\% WER in user response recognition. Ablation studies confirm significant gains in specific entity recognition: hotel names (30.6\% vs 42.1\% SER), train destination stations (39.6\% vs 63.1\%SER), and restaurant names (30.9\% vs 40.1\%SER).

While our experiments in this paper focus on the dialog state tracking task, ReSLM's applicability extends beyond. For example, ReSLM can be used for contextual ASR with biasing functions as described in the SLM paper~\cite{SLM2023}.

\section{Acknowledgements}
We would like to acknowledge Jeffrey Zhao, Zelin Wu, Gang Li, Yongqiang Wang and other colleagues for useful discussions and pointers.

\bibliographystyle{IEEEtran}
\bibliography{mybib}

\end{document}